\documentclass{article}
\usepackage{spconf,amsmath,graphicx}

\usepackage{amsmath,amsfonts,bm}

\def\eqref#1{equation~\ref{#1}}

\def\1{\bm{1}}

\DeclareMathAlphabet{\mathsfit}{\encodingdefault}{\sfdefault}{m}{sl}
\SetMathAlphabet{\mathsfit}{bold}{\encodingdefault}{\sfdefault}{bx}{n}

\DeclareMathOperator*{\argmax}{arg\,max}

\usepackage{hyperref}
\hypersetup{bookmarks=false,colorlinks}

\usepackage{xurl}

\usepackage{algorithm} 
\usepackage{algpseudocode}
\usepackage{multirow}

\algrenewcommand\algorithmicrequire{\textbf{Input:}}
\algrenewcommand\algorithmicensure{\textbf{Output:}}

\usepackage[outdir=./]{epstopdf}
\usepackage{enumitem}

\def\@name{ \emph{A. N. Author},  \\ \emph{A. N. Other Author}}

\title{Learning with Instance-Dependent Noisy Labels by Anchor Hallucination and Hard Sample Label Correction}

\name{Po-Hsuan Huang$^{1,\star}$ \quad Chia-Ching Lin$^{1,\star}$ \quad Chih-Fan Hsu$^{1}$ \quad
Ming-Ching Chang$^{2}$ \quad  Wei-Chao Chen$^{1}$ \thanks{$^{\star}$Equal contribution}}
\address{$^{1}$Inventec Corporation, Taipei, Taiwan \\
      $^{2}$Department of Computer Science, University at Albany, State University of New York, NY, USA}
      
\begin{document}
%
\maketitle
%
%
\begin{abstract}

Learning from noisy-labeled data is crucial for real-world applications. Traditional Noisy-Label Learning (NLL) methods categorize training data into clean and noisy sets based on the loss distribution of training samples. 
However, they often neglect that clean samples, especially those with intricate visual patterns, may also yield substantial losses. This oversight is particularly significant in datasets with Instance-Dependent Noise (IDN), where mislabeling probabilities correlate with visual appearance.
Our approach explicitly distinguishes between clean {\em vs.} noisy and easy {\em vs.} hard samples. We identify training samples with small losses, assuming they have simple patterns and correct labels. Utilizing these easy samples, we hallucinate multiple anchors to select hard samples for label correction. 
Corrected hard samples, along with the easy samples, are used as labeled data in subsequent semi-supervised training.
Experiments on synthetic and real-world IDN datasets demonstrate the superior performance of our method over other state-of-the-art NLL methods.

\end{abstract}

\begin{keywords}
Noisy-label learning, instance-dependent noise, anchor hallucination, semi-supervised learning
\end{keywords}

\section{Introduction}
\label{sec:intro}
\vspace{-1mm}

The success of Deep Neural Networks (DNNs) heavily relies on extensively annotated datasets. However, data annotation is often costly and inevitably comes with label noise~\cite{xiao15clothing1m, wei2021cifar10N}. The correction of label errors and the exploration of robust representations have become focal points in recent research~\cite{song22survey}. In this paper, we consider practical real-world scenarios of training an image classification model from a dataset with noisy labels~\cite{wei2021cifar10N}, where the probability of mislabeling each image is contingent on its visual appearance, characterized as Instance-Dependent Noise (IDN).

\begin{figure}[t]
\centerline{
\includegraphics[width=\linewidth]{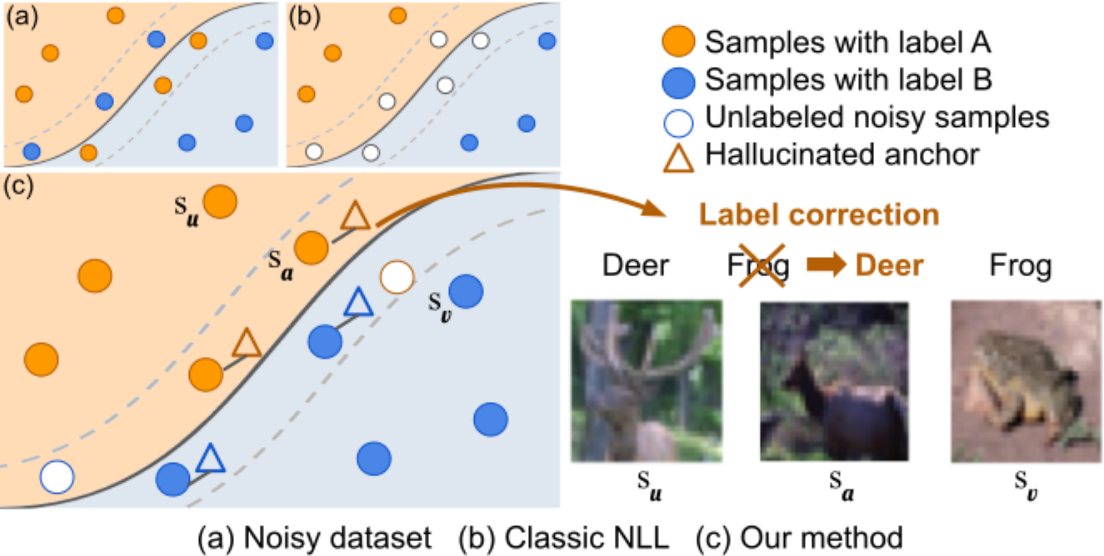}
\vspace{-2mm}
}
\caption{
{\bf A schematic plot of our method for visual classification in comparison with classic NLL methods.}
(a) Classification on a noisy dataset with Instance-Dependent Noise (IDN) noisy labels. (b) Existing selection-based NLL methods~\cite{li20dividemix,karim21unicon} treat large-loss samples near the decision boundary that are hard to classify as unlabeled data. (c) Our proposed method identifies the hard samples and corrects their labels through anchor hallucination and selection.
}
\vspace{-4mm}
\label{fig:hard_samples_idn}
\end{figure}

Traditional Noisy-Label Learning (NLL) methods primarily rely on \textit{sample selection}~\cite{li20dividemix, karim21unicon}. These methods identify correctly-labeled ({\em i.e.}, {\em clean}) samples in the training set by employing the {\em small-loss} criterion during an initial training. Samples with small classification loss are assumed to have correct labels, and samples with large loss have potentially incorrect labels.
However, there is a major drawback to this approach.
DNNs are known to prioritize learning simple patterns over complex ones~\cite{Arpit17}. Consequently, the initially identified small-loss samples may only represent a subset of clean samples with simple visual patterns. Conversely, labels of samples with large classification losses are not necessarily always \textit{noisy}; they may have clean labels that are {\em hard} to learn due to their complex visual patterns.
For example, in CIFAR-10~\cite{Krizhevsky09}, airplanes are typically in the sky, and ships are usually on the water; but a few airplane samples also appear on the water that are harder to classify. We argue that these hard samples with clean labels are important to characterize the decision boundary for effective model learning. Simply discarding their potentially noisy labels, as common in typical NLL methods, results in performance degradation~\cite{chen2021beyond}.

\begin{figure*}[t]
\centerline{
\includegraphics[width=0.9\textwidth]{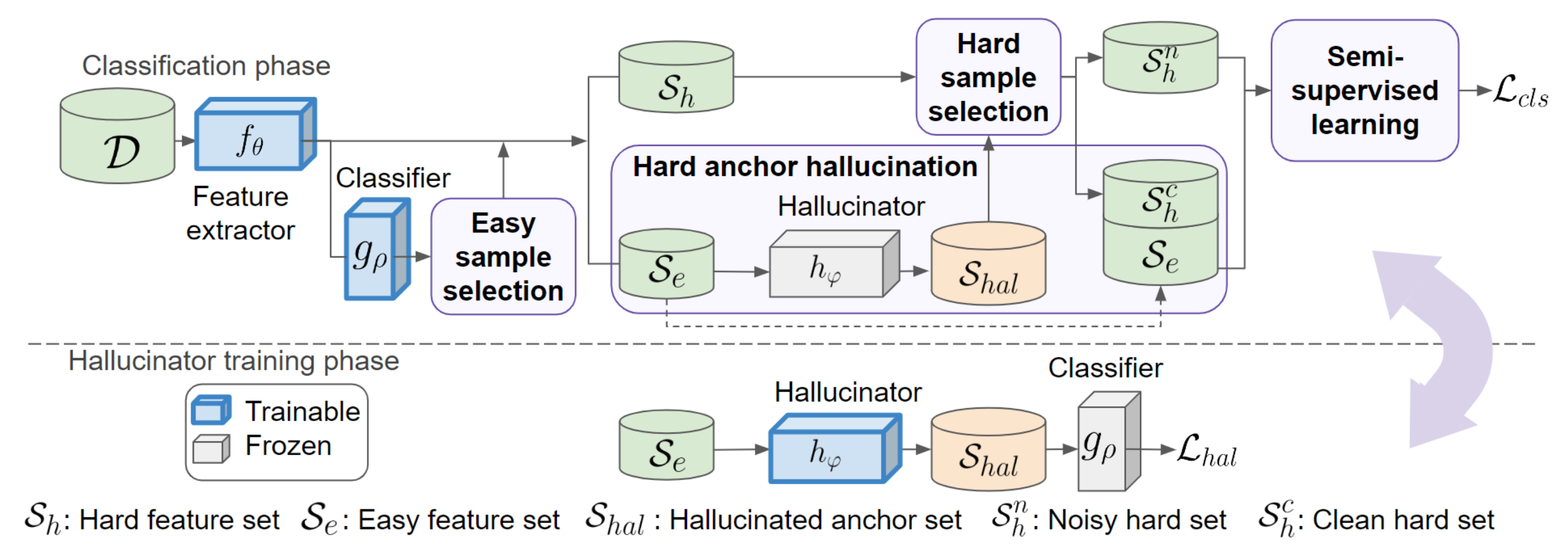}
\vspace{-2mm}
}
\caption{{\bf Our NLL learning framework} consists of two main training phases, namely the {\em classification} phase and the {\em hallucinator training} phase. The classification phase consists of four steps: (1) easy sample selection, (2) hard anchor hallucination, (3) hard sample selection, and (4) semi-supervised learning. The hallucinator model is updated in the hallucinator training phase.
}
\vspace{-4mm}
\label{fig:framework}
\end{figure*}

To address this challenge, our approach distinguishes {\em easy} from {\em hard} samples, in addition to differentiating between samples with {\em clean} {\em vs.} {\em noisy} labels. 
Fig.~\ref{fig:hard_samples_idn} illustrates our method, starting with easy samples to identify and correct labels of hard samples. Initially, we use the standard small-loss criterion to partition the training set into class-balanced easy and hard samples. To identify hard samples for label correction, we introduce a novel {\bf anchor hallucination} technique that synthesizes feature vectors in the feature space. These hallucinated features, called {\bf anchors}, are generated from easy samples to simulate hard samples that are with complex visual compositions. Anchors are then used to select hard samples from the original training set in the nearby feature space, see Fig.~\ref{fig:hard_samples_idn}(c). Since anchors are synthesized with pre-known labels, they are used to correct the labels of the selected hard samples via majority voting. Lastly, the label-corrected hard samples together with the initial easy samples constitute the labeled data, while the remaining samples are used as unlabeled data in a semi-supervised model training. This effectively enhances the data and label utilization.

Contributions of this paper are summarized as follows:
\begin{itemize}[leftmargin=10pt] \itemsep -.01em

\item We focus on the Noisy-Label Learning (NLL) problem, in particular data with Instance-Dependent Noise (IDN). We advocate for differentiating easy and hard training samples, while also distinguishing their clean {\em vs.} noisy labels. This design address the sample selection dilemma in NLL.

\item We introduce an anchor hallucination technique by synthesizing feature vectors for selecting hard samples and performing label correction for them. Followed by a semi-supervised model training to maximize data and label utilization.

\item Extensive experiments are conducted on synthetic IDN datasets derived from CIFAR-10~\cite{Krizhevsky09}, as well as real-world CIFAR-10N/100N~\cite{wei2021cifar10N} and Clothing1M~\cite{xiao15clothing1m} datasets.
The results demonstrate the superiority of our method over the State-of-The-Art (SoTA) NLL methods, including DivideMix~\cite{li20dividemix} and TSCSI~\cite{zhao22tscsi}.

\end{itemize}

\section{Related works}
\label{sec:relatedwork}
\vspace{-1mm}

{\bf Noisy-Label Learning (NLL).}
In NLL literature, Instance-Independent Noise (IIN) stands out as the most prevalent type of label noise. IIN is characterized by the probability of an image being mislabeled, which depends solely on the involved class pair, regardless of its visual content. Notable examples of IIN include symmetric and asymmetric noise patterns proposed in \cite{Patrini17}, which have gained widespread adoption in related fields. Various NLL approaches are developed based on this noise assumption, encompassing the design of noise-robust loss functions~\cite{ma2020normalized}, loss correction~\cite{Wang20}, label correction~\cite{wang2021proselflc}, and sample selection~\cite{li20dividemix}.

Recent prominent studies regarding IIN combine sample selection with Semi-Supervised Learning (SSL), yielding notable progress~\cite{li20dividemix,karim21unicon}. The majority of methods resort to the small-loss criterion and consider samples with small training loss as clean. Subsequently, an off-the-shelf SSL algorithm, such as~\cite{berthelot2019mixmatch}, is applied, treating the selected samples as labeled data and the remainder as unlabeled. However, these approaches often overfit to a small training subset of easy samples chosen based on the small-loss criterion~\cite{chen2021beyond}. This limitation hampers their ability to fully exploit critical labeling information contained in hard samples near the decision boundary. Despite their success on various IIN benchmarks, performance under the IDN assumption remains unclear.

{\bf Learning from Instance-Dependent Noise (IDN) Data.}
In contrast to the naive assumption of IIN, recent works~\cite{chen2021beyond,xia2020part,Zhang21PMD,wei2021cifar10N,zhu2021CAL,cheng2022MEIDTM,zhao22tscsi} contend that real-world noise patterns are more likely to depend on visual content, prompting a shift towards addressing IDN. 
Some methods combat IDN by estimating the noise transition matrix~\cite{cheng2022MEIDTM,berthon2021confidence}, requiring additional information and achieving mediocre performance on real-world data. Others adopt a selection-based method combined with SSL, similar to previous works for IIN~\cite{zhao22tscsi}, and have reached SoTA results on several IDN benchmarks. However, the effective utilization of valuable hard samples with potentially noisy labels remain an unsolved challenge. Our work aligns with this research line, focusing on leveraging robust information from the easy samples to: (1) identify hard samples with noisy labels, and (2) perform label correction.

\section{The proposed method}
\label{sec:method}
\vspace{-1mm}

This paper addresses the noisy label learning of image classification for data with Instance-Dependent Noise (IDN). We work with a noisy training set $\mathcal{D}={(x_n,\tilde{y}_n)}$, where $x_n$ denotes the $n$-th image, and $\tilde{y}_n \in \{ 1,2,...,C\}$ is the corresponding label for $C$ classes. The given label $\tilde{y}_n$ may differ from the real ground truth label $y_n$, which remains unobservable during training. The goal is to train an image classification model on $\mathcal{D}$ that performs well on a clean test set.

Our approach operates by distinguishing between easy and hard training samples, using the cleanly-labeled easy samples to identify hard samples for label correction. This process is achieved through {\bf anchor hallucination}, where features are synthesized from the easy samples to create anchors. These anchors are then employed to select hard samples via majority voting for label correction.
Overall, our model is structured with a classification module, comprising a feature extractor $f_{\theta}$ and a linear classifier $g_{\rho}$, along with an anchor hallucinator $h_{\phi}$.

Our training framework comprises two iterative phases, as in Fig.~\ref{fig:framework}.
First, in the {\bf classification} phase, we keep  $h_{\phi}$ fixed and optimize $f_{\theta}$ and $g_{\rho}$ via semi-supervised training ($\S$~\ref{ssec:ssl}) using clean-labeled easy sample and label-corrected hard samples. The pipeline comprises three steps: easy sample selection ($\S$~\ref{sec:easy:sample:sel}), hard anchor hallucination ($\S$~\ref{ssec:hard_hal}), and hard sample selection ($\S$~\ref{ssec:hard_ret}). 
Secondly, in the {\bf hallucinator training} phase, $f_{\theta}$ and $g_{\rho}$ remain fixed, and only $h_{\phi}$ undergoes updates based on a hallucination loss outlined in $\S$~\ref{ssec:hard_hal}.
Further details regarding the steps and losses are elaborated in the subsequent sections.

\vspace{-2mm}
\subsection{Easy sample selection}
\label{sec:easy:sample:sel}
\vspace{-1mm}

Our method starts with selecting {\em easy} samples using small-loss criterion~\cite{li20dividemix}. Drawing on the insight that DNNs tend to learn simple patterns faster than complex ones~\cite{Arpit17}, we identify easy samples by analyzing the distribution of classification losses during initial training. 
Specifically, we compute the cross-entropy loss for each sample $(x_n,\tilde{y}_n) \in \mathcal{D}$ and fit the loss distribution across all training samples using a two-component Gaussian Mixture Model (GMM), which provides greater flexibility in capturing sharpness of the fitted distribution \cite{li20dividemix}. 
The Gaussian component with the smaller mean represents the distribution of smaller losses. The probability of each sample belonging to this Gaussian component is then calculated as the {\em easiness score} $\omega_{n}$ for that sample.

Subsequently, we select a fixed fraction of samples with the top-$P\%$ easiness scores to form the set of {\em easy} training samples.
The hyperparameter $P$ defaults to 60 and is fine-tuned using a small clean validation set. To maintain balance across all classes, we ensure a sufficient number of easy samples for each class.
Let $N_j$ denote the total number of samples of the $j$-th class in the training set $\mathcal{D}$. We control the number of easy samples $M_{j}$ for the $j$-th class to be:
\begin{equation}
M_{j}= \min
\left(
\left\lceil |\mathcal{D}| \times 
\frac{P\%}{C}
\right\rceil, 
N_{j}
\right).
\label{eq1:easy_selection}
\end{equation}
We thus obtain class-balanced easy training samples with assumed clean labels. The remaining samples with potentially noisy labels are regarded as hard samples. We next utilize the feature extractor $f_{\theta}$ to embed both easy and hard samples into a $d$-dimensional feature space, resulting in the \textit{easy feature set} $\mathcal{S}_{e}$ and {\em hard feature set} $\mathcal{S}_{h}$, which will be used for anchor hallucination, hard sample selection and label correction.

\vspace{-2mm}
\subsection{Hard anchor hallucination}
\label{ssec:hard_hal}
\vspace{-1mm}

The easy feature set $\mathcal{S}_{e}$ comprises feature vectors of training samples with simple visual patterns and clean labels. By combining features from $\mathcal{S}_{e}$, we can form complex visual patterns, which form the basis for hard anchor hallucination. Through feature concatenation and generating a substantial number of feature anchors spanning the feature space, we can employ majority voting of nearby anchors to search for the most matching feature vectors in the hard feature set $\mathcal{S}_{h}$.

Anchors are hallucinated by aggregating features of easy samples in an automatic, data-driven process. Specifically, for each $s_u \in \mathcal{S}_{e}$ from class $\tilde{y}_u$, we randomly select another feature $s_v \in \mathcal{S}_{e}$ from a different class $\tilde{y}_v$ (where $\tilde{y}_v \ne \tilde{y}_u$) for mixing. Subsequently, we concatenate $s_u$ and $s_v$, and feed it to the hallucinator $h_{\phi}$ to produce a hallucinated anchor, $s_a = h_{\phi}(s_u, s_v)$.
To ensure that $s_a$ is transformed into a hard anchor of the desired class, we formulate the {\em hallucination loss} $L_{hal}$ according to the following two designs.

First, to encourage the hallucinated anchor $s_a$ represents a \textit{hard} instance, we optimize the hallucinator $h_{\phi}$ by regularizing the \textit{similarities} between $s_a$ and both $s_u$ and $s_v$.
Specifically, we define a {\em similarity loss} based on the cosine distances between features as $L_{sim}=-\lambda_{p} \left\langle s_a, s_u \right\rangle -(1-\lambda_p) \left\langle s_a, s_v \right\rangle$, where $\lambda_p \in [0.5, 1.0]$ is a hyperparameter controlling the difficulty level of $s_a$, and $\left\langle \cdot, \cdot \right\rangle$ computes the cosine similarity between its arguments. By minimizing $L_{sim}$, the hallucinated anchor $s_a$ will be encouraged to reside in the area between $s_u$ and $s_v$ in the feature space, and thus share visual patterns from both classes $\tilde{y}_v$ and $\tilde{y}_u$.

Second, to ensure hallucinated anchor $s_a$ belongs to a known, desired class, we follow the work of \cite{zhang2021hallucination} and define a classification loss using its target label $\tilde{y}_a=\tilde{y}_u$.
The overall hallucination loss $L_{hal}$ is calculated as:
\begin{equation}\label{eq2:loss_hal}
L_{hal} = L_{sim} + \mathcal{H}(s_a, \tilde{y}_u),
\end{equation}
where $\mathcal{H}(\cdot,\cdot)$ computes the cross-entropy loss. 
Minimizing Eq.~(\ref{eq2:loss_hal}) prompts the hallucinator to generate an anchor $s_a = h_{\phi}(s_u, s_v)$ with complex visual patterns positioned near the decision boundary between classes $\tilde{y}_u$ and $\tilde{y}_v$, while remaining on the side closer to $\tilde{y}_u$. Multiple hallucinated anchors can be produced from a single feature $s_u$ by sampling different $s_v$. The set of hallucinated anchors is denoted as $\mathcal{S}_{hal}$.

\vspace{-2mm}
\subsection{Hard sample selection}
\label{ssec:hard_ret}
\vspace{-1mm}

The hallucinated anchors extend throughout the feature space, with varying degrees of \textit{representative} qualities.
Computationally, a hallucinated anchor $s_a \in \mathcal{S}_{hal}$ is considered representative of a real hard feature sample $s_h \in \mathcal{S}_{h}$, if they are sufficiently close in the feature space.
We measure this proximity between $s_a$ and $s_h$ using cosine similarity $\left\langle \cdot , \cdot \right\rangle$. To filter out the representative $s_a$, we identify the nearest hard feature sample of a hallucinated anchor $s_a$, denoted as $s_r = \argmax_{s_n \in \mathcal{S}_{h}} \left\langle s_a, s_n\right\rangle$. The anchor $s_a$ is considered a \textit{valid} representative of $s_r$, if $\left\langle s_a, s_r\right\rangle$ exceeds a threshold $\lambda_{conf}$.
The hyperparameter $\lambda_{conf}$ can be tuned based on a small validation set with clean labels. These valid representatives $s_a$ disperse near the hard feature samples and can be used to match them. 
For each $s_h \in \mathcal{S}_{h}$, we collect up to $K$ of its surrounding valid representatives. These $K$ valid representatives are leveraged to determine the correct label of $s_h$ by majority voting.

By identifying hard samples with corrected labels in this manner, the classification module ($f_{\theta}$ and $g_{\rho}$) is trained on more valuable labeling information contained in the hard samples, leading to improved performance.

\vspace{-2mm}
\subsection{Semi-supervised learning}
\label{ssec:ssl}
\vspace{-1mm}

The set of hard samples with corrected labels, denoted $\mathcal{S}^{c}_{h}$, is combined with the easy feature set $\mathcal{S}_{e}$ to form the labeled dataset $\mathcal{S}_{labeled} = \mathcal{S}^{c}_{h} \cup \mathcal{S}_{e}$ for model training, while the remaining noisy hard features, denoted as $\mathcal{S}^{n}_{h}$, constitute the unlabeled dataset $\mathcal{S}_{unlabeled} = \mathcal{S}^{n}_{h}$ for semi-supervised learning (SSL).

We adopt the classic SSL method MixMatch~\cite{berthelot2019mixmatch} to augment samples from $\mathcal{S}_{labeled}$ and $\mathcal{S}_{unlabeled}$.
We obtain pseudo labels for samples in $\mathcal{S}_{unlabeled}$ from the average of model predictions across its two augmented copies. Similarly, we refine the labels of samples in $\mathcal{S}_{labeled}$ by using a linear combination of their given labels in $\mathcal{D}$ and the average of model predictions across the two augmented images. The weight assigned to the given label is determined by the easiness score $\omega_{n}$ from $\S$~\ref{sec:easy:sample:sel}.

After obtaining the refined pseudo labels, we then perform SSL using the combined classification loss:
%
\begin{equation}
\mathcal{L}_{SSL}= \mathcal{L}_{CE}+\lambda_{MSE} \; \mathcal{L}_{MSE},
\label{eq5:loss_ssl}
\end{equation}
where $\mathcal{L}_{CE}$ is the cross-entropy loss for the labeled data, $\mathcal{L}_{MSE}$ is the mean squared error for the unlabeled data, and $\lambda_{MSE}$ is a hyperparameter set through validation.
By minimizing Eq.~(\ref{eq5:loss_ssl}), the resulting image classifier 
comprising feature extractor
$f_{\theta}$ and linear classifier $g_{\rho}$ would become more robust, as it incorporates information from critical hard samples with clean labels during training.

\vspace{-2mm}
\subsection{Iterative model training}
\vspace{-1mm}

To prevent the hallucinator $h_{\phi}$ from degeneration, i.e., always producing identical hallucinated anchors $s_a$ regardless of the input pair $(s_u, s_v)$, we adopt an iterative training procedure, as illustrated in Fig.~\ref{fig:framework}. After a few epochs of warm-up training, we start the iterative training stage, which consists of two training phases. In the \textit{classification} phase, we freeze the hallucinator $h_{\phi}$ and train the feature extractor $f_{\theta}$ and the linear classifier $g_{\rho}$ jointly using the corrected labeled and unlabeled training sets ($\mathcal{S}_{labeled}$ and $\mathcal{S}_{unlabeled}$) 
in $\S$~\ref{ssec:ssl} by minimizing Eq.~(\ref{eq5:loss_ssl}).
In the \textit{hallucinator training} phase, we freeze $f_{\theta}$ and $g_{\rho}$ and train $h_{\phi}$ by minimizing Eq.~(\ref{eq2:loss_hal}). The two phases are performed iteratively until sufficient epochs are reached.

\section{Experiments}
\vspace{-1mm}


\subsection{Datasets and IDN noise generation}
\vspace{-2mm}

We follow previous NLL works on learning from datasets with IDN labels~\cite{xia2020part,chen2021beyond,zhu2021CAL,zhao22tscsi} to conduct the experiments on both synthetic and real-world IDN datasets.

{\bf Synthetic IDN datasets.}
We conduct experiments on synthetic IDN datasets created from the CIFAR-10 dataset~\cite{Krizhevsky09}, which contains 50,000 training images and 10,000 test images from 10 cleanly annotated classes.
We considered two approaches in generating IDN noises: 1) \textit{part-dependent label noise} (PTD)~\cite{xia2020part}, which is generated according to a combination of multiple noise transition matrices of different \textit{parts} of an image; 2) \textit{classification-based label noise}~\cite{chen2021beyond}, which is generated by averaging the collected softmax outputs during training using a standard CNN trained on all the training data for multiple epochs.

{\bf Real-world IDN datasets.}
To evaluate the effectiveness of our method on real-world IDN datasets, we conducted experiments using the CIFAR-10N/100N~\cite{wei2021cifar10N} and the real-world Clothing1M~\cite{xiao15clothing1m} datasets. CIFAR-10N/100N were generated from CIFAR-10/100 by collecting labels from three human annotations for each training image through Amazon Mechanical Turk. The three noisy labels for each image are denoted as \textit{Random 1/2/3}, and are further aggregated by majority vote (denoted as \textit{Aggregate}) and by random selection of one wrong label if there is any (denoted as \textit{Worst}). The Clothing1M dataset contains over 1 million training images of 14 different types of clothing collected online, with labels extracted from the surrounding text of images. We use the 14K clean validation set for hyperparameter tuning and the 10K clean test set to evaluate the model performance. These IDN datasets present real-world scenarios with various noise sources and thus provide a suitable testbed for comparing our method with the SoTA methods.

\vspace{-2mm}
\subsection{Baselines and implementation details}
\vspace{-1mm}

We compare our framework with recent SoTA NLL works, including those focusing on IIN datasets such as DivideMix~\cite{li20dividemix}, and those focusing on IDN datasets such as TSCSI~\cite{zhao22tscsi}. It is worth noting that both DivideMix and TSCSI employ two networks in a co-training fashion for model ensemble thus incurring higher computational costs, whereas our framework only trains a single network in most of our experiment settings except on Clothing1M.
For CIFAR-10 with IDN and the CIFAR-10N/100N datasets, we follow previous works~\cite{wei2021cifar10N,zhao22tscsi} and adopt ResNet-34 network as our classification module, and a two-layer Multi-Layer Perceptron (MLP) as our hallucinator $h_{\phi}$.
We evaluate our method on a clean testing set and report the best testing accuracy on average over three runs.
As for Clothing1M, we adopt an ImageNet-pretrained ResNet-50 network as per the prior works~\cite{li20dividemix,zhao22tscsi} while also implementing $h_{\phi}$ as a two-layer MLP. We also adopt the same procedures as those used in DivideMix to select easy samples(GMM-based selection without class balancing) for better comparison.
During training, we use the 14K clean validation set to choose the best model, which is applied to the 10K clean test to get the test accuracy.
More implementation details are available in the supplementary materials.

\vspace{-3mm}
\subsection{Quantitative results}
\vspace{-2mm}

{\bf Results on PTD label noise.}
Table~\ref{table:cifar10_ptd} shows experimental results on the CIFAR-10 datasets with PTD noise~\cite{xia2020part}. Our proposed method achieves significant performance improvement compared to prior state-of-the-art methods under both 20\% and 40\% noise ratios. Our model also shows robustness against the increasing noise rate under PTD.

\begin{table}[t]
  \caption{Classification accuracy (\%) on CIFAR-10 with ParT-Dependent (PTD) label noise~\cite{xia2020part} of 20\% and 40\%. Scores of the baseline methods are taken from \cite{zhao22tscsi}. Best results are in \textbf{bold} and the second best are \underline{underlined}.
  \vspace{1mm}
  }
  \label{table:cifar10_ptd}
\centerline{
  \small
    \begin{tabular}{lcc}
    \hline
    Method & PTD 20\% & PTD 40\%  \\
    \hline

    \multirow{1}{*}{Forward T~\cite{Patrini17}}  & 87.22$\pm$1.60 & 79.37$\pm$2.72 \\
  


    \multirow{1}{*}{Co-teaching~\cite{Han18}}  & 88.87$\pm$0.24 & 73.00$\pm$1.24 \\
  
    \multirow{1}{*}{Co-teaching+~\cite{Yu19}}  & 89.80$\pm$0.28 & 73.78$\pm$1.39\\
    
    \multirow{1}{*}{JoCoR~\cite{wei2020combating}}  & 88.78$\pm$0.15 & 71.64$\pm$3.09  \\
 



    \multirow{1}{*}{DivideMix~\cite{li20dividemix}}  & 93.33$\pm$0.14 & \underline{95.07}$\pm$0.11 \\

    \multirow{1}{*}{CAL~\cite{zhu2021CAL}}  & 92.01$\pm$0.75 &    84.96$\pm$1.25 \\
 
    \multirow{1}{*}{TSCSI~\cite{zhao22tscsi}}  & \underline{93.68}$\pm$0.12 & 94.97$\pm$0.09  \\

    \multirow{1}{*}{Ours}  & \textbf{94.26}$\pm$0.19 & \textbf{95.28}$\pm$0.10 \\

    \hline
  \end{tabular}
}
\vspace{-4mm}
\end{table}

\begin{table}[t]
\caption{Classification accuracy (\%) on CIFAR-10 with classification-based label noise~\cite{chen2021beyond} across different noise ratios. Results of the baseline methods are taken from \cite{zhao22tscsi}.
\vspace{1mm}
}
\label{table:cifar10_idn}
\centerline{
\setlength{\tabcolsep}{0.9mm}
  \small
  \begin{tabular}{lccccc}
    \hline
    Method & 10\% & 20\% &40\% \\
    \hline
    \multirow{1}{*}{CE (Standard)}  & 91.25$\pm$0.27& 86.34$\pm$0.11  & 75.68$\pm$0.29 \\
    \multirow{1}{*}{Forward~\cite{Patrini17}}  & 91.06$\pm$0.02& 86.35$\pm$0.11  & 71.12$\pm$0.47 \\
    \multirow{1}{*}{Co-teaching~\cite{Han18}}  & 91.22$\pm$0.25& 87.28$\pm$0.20  & 78.82$\pm$0.47 \\

    \multirow{1}{*}{DAC~\cite{thulasidasan2019combating}}  & 90.94$\pm$0.09 & 86.16$\pm$0.13 & 74.80$\pm$0.32 \\

    \multirow{1}{*}{SEAL~\cite{chen2021beyond}}  & 91.32$\pm$0.14& 87.79$\pm$0.09 & 82.98$\pm$0.05 \\
    \multirow{1}{*}{TSCSI~\cite{zhao22tscsi}}  & \underline{91.39}$\pm$0.08 & \underline{88.36}$\pm$0.11& \underline{84.18}$\pm$0.40  \\
    \multirow{1}{*}{Ours}  & \textbf{93.68}$\pm$0.47& \textbf{92.98}$\pm$0.11 & \textbf{92.47}$\pm$0.41 \\
    \hline
  \end{tabular}
}
\vspace{-4mm}
\end{table}

\begin{table*}[t]
  \caption{Classification accuracy (\%) on CIFAR-10N/100N~\cite{wei2021cifar10N} across different noise settings. Baseline scores are from \cite{wei2021cifar10N}. 
\vspace{1mm}
  }
\label{table:cifar_n}
\centerline{
  \small
    \begin{tabular}{lccccc}
    \hline
    \multicolumn{1}{l}{}
    & \multicolumn{4}{c}{CIFAR-10N}
    & \multicolumn{1}{c}{CIFAR-100N} \\
    Method & Random1 &Random2 &Random3 &Worst & Noisy  \\
    \hline
    \multirow{1}{*}{CE (Standard)} & 85.02$\pm$0.65&86.46$\pm$1.79 &85.16$\pm$0.61  & 77.69$\pm$1.55 & 55.50$\pm$0.66 \\
    \multirow{1}{*}{Forward T~\cite{Patrini17}} & 86.88$\pm$0.50 &  86.14$\pm$0.24  &  87.04$\pm$0.35  &  79.79$\pm$0.46  &  57.01$\pm$1.03 \\
    \multirow{1}{*}{Co-teaching+~\cite{Yu19}} &89.70$\pm$0.27  &89.47$\pm$0.18 &89.54$\pm$0.22  & 83.26$\pm$0.17  & 57.88$\pm$0.24 \\

    \multirow{1}{*}{DivideMix~\cite{li20dividemix}} &\underline{95.16}$\pm$0.19 &\underline{95.23}$\pm$0.07 &95.21$\pm$0.14  & 92.56$\pm$0.42 & \textbf{71.13}$\pm$0.48 \\
    \multirow{1}{*}{Negative-LS ~\cite{Wei2022ToSO}} &90.29$\pm$0.32 &90.37$\pm$0.12 &90.13$\pm$0.19  &   82.99$\pm$0.36   & 58.59$\pm$0.98 \\
    \multirow{1}{*}{VolMinNet ~\cite{li2021provably}} &88.30$\pm$0.12 &88.27$\pm$0.09 &88.19$\pm$0.41 & 80.53$\pm$0.20  & 57.80$\pm$0.31 \\
    \multirow{1}{*}{CAL~\cite{zhu2021CAL}} &90.93$\pm$0.31 &90.75$\pm$0.30 &90.74$\pm$0.24  & 85.36$\pm$0.16 & 61.73$\pm$0.42 \\
    \multirow{1}{*}{PES~\cite{bai2021PES}} &95.06$\pm$0.15 &95.19$\pm$0.23 &\underline{95.22}$\pm$0.13 & \underline{92.68}$\pm$0.22 &  70.36$\pm$0.33 \\
    \multirow{1}{*}{Ours}  &\textbf{95.21}$\pm$0.05 &\textbf{95.31}$\pm$0.10  &\textbf{95.25}$\pm$0.17   & \textbf{93.52}$\pm$0.49 &  \underline{70.79}$\pm$0.06  \\
    \hline
  \end{tabular}
}
\end{table*}


{\bf Results on classification-based label noise.}
Table~\ref{table:cifar10_idn} provides performance comparisons on CIFAR-10 with classification-based label noise~\cite{chen2021beyond} under different noise levels. The classification-based label noise is considered challenging due to its originating from a classification model~\cite{zhao22tscsi}. Our method consistently demonstrates significantly superior performance compared to previous methods across all noise levels. Our method exhibits remarkable resistance to higher levels of label noise (40\%) on classification-based label noise, while other methods suffer substantial performance degradation.

\begin{table}[t]
  \caption{Classification accuracy (\%) on Clothing1M. We report our baseline DivideMix on average over three runs using their official code. Results of other methods are from \cite{zhao22tscsi}. 
\vspace{1mm}  
}
  \label{table:c1m}
\centerline{
\resizebox{\columnwidth}{!}{%
\setlength{\tabcolsep}{0.5mm}
  \scriptsize
  \begin{tabular}{lccccccc}
    \hline
   Method & Co-teaching & JoCoR &DivideMix &  CAL &  TSCSI & Ours  \\
          &  ~\cite{Han18}&~\cite{wei2020combating}&~\cite{li20dividemix}&~\cite{zhu2021CAL}&~\cite{zhao22tscsi}&\\
   \hline
   Accuracy & 69.21 &70.30 & 74.40$\pm$0.08 & 74.17& \textbf{75.40} & \underline{74.62$\pm$0.14}\\

    \hline
  \end{tabular} 
  }
}
\vspace{-4mm}
\end{table}

\begin{table}[t!]
\caption{Ablation analysis on CIFAR-10 with 40\% classification based noise~\cite{chen2021beyond}.
\vspace{1mm}
}
\label{table:ablation}
\centerline{
  \small
  \begin{tabular}{ccc}
    \hline
    Easy sample sel. & Hard sample corr. & Test accuracy \\
    \hline
    - & - & 86.24$\pm$0.90 \\
    $\checkmark$ & - &89.77$\pm$1.45  \\
    $\checkmark$ & $\checkmark$ & \textbf{92.47}$\pm$0.41 \\
    \hline
  \end{tabular}
}
\vspace{-5mm}
\end{table}


{\bf Results on CIFAR-N.} Table~\ref{table:cifar_n} shows performance comparisons on the CIFAR-10N/CIFAR-100N datasets~\cite{wei2021cifar10N}. Our method consistently outperforms other methods on CIFAR-10N with all the noise settings of \textit{Random 1, Random 2, Random 3}, and \textit{Worst}.
Notably, our method achieves comparable performance against DivideMix~\cite{li20dividemix} on CIFAR-100N while only trained a single network.
This demonstrates the efficacy of our method in learning from real-world IDN datasets.

\begin{figure*}[t]
\centerline{
\includegraphics[width=0.70\linewidth]{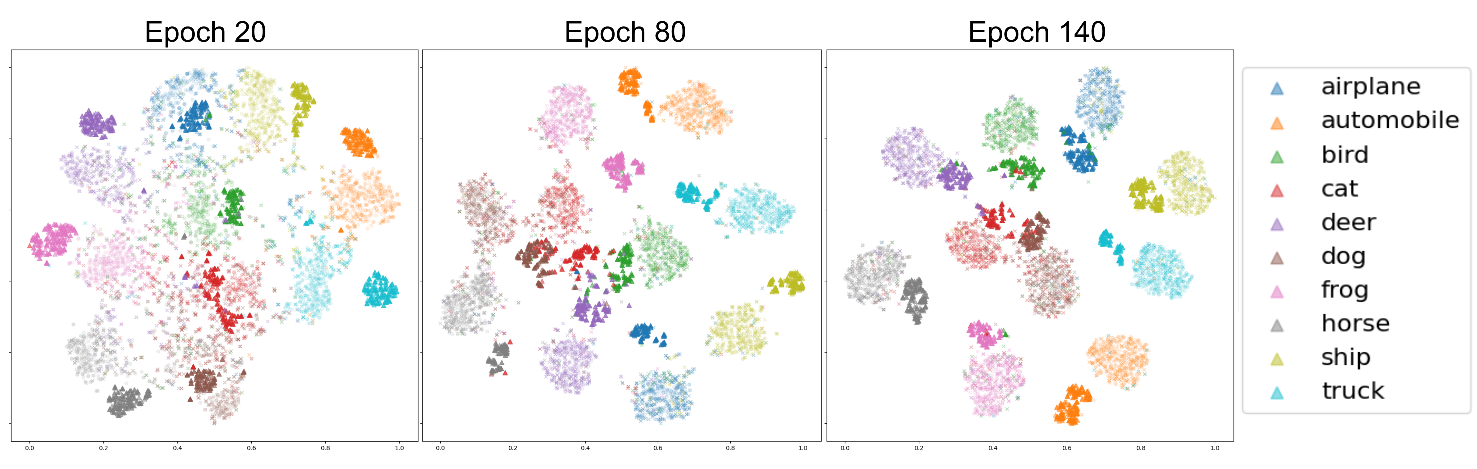}
\vspace{-4mm}
}
\caption{
{\bf The t-SNE visualization for the hallucinated anchors.} 
We use darker colors to denote the hallucinated anchors and lighter colors for real samples. Note that most of the hallucinated anchors are distributed around the decision boundary.}
\vspace{-3mm}
\label{fig:exp:tsne}
\end{figure*}

{\bf Results on Clothing1M.}
Table~\ref{table:c1m} shows performance comparisons on the Clothing1M dataset. Our method achieves competitive results compared with TSCSI and is superior to DivideMix and other methods. Since our method adopts similar strategies with DivideMix in easy sample selection ($\S$~\ref{sec:easy:sample:sel}), the superior performance compared to DivideMix indicates the effectiveness of our hallucination-based hard sample selection ($\S$~\ref{ssec:hard_hal} and $\S$~\ref{ssec:hard_ret}) in learning from such a large-scale IDN dataset.

{\bf Ablation study.}
To evaluate the effectiveness of each design component, we conducted an ablation analysis of our proposed framework on the CIFAR-10 dataset with 40\% classification-based IDN. We compared the performance of three different settings: (1) vanilla GMM selection-based method, which is essentially DivideMix~\cite{li20dividemix} without co-training and model ensemble, (2) our method with only the easy sample selection stage as described in $\S$~\ref{sec:easy:sample:sel}, and (3) our method with both stages of easy sample selection and hard sample correction in $\S$~\ref{ssec:hard_ret}. Table~\ref{table:ablation} presents the comparison results. As can be seen from the table, the design of each of the two stages contributes to the performance improvement of our framework. Notably, the easy sample selection stage contributed the most to the performance boost, indicating the importance of obtaining a class-balanced easy subset for effective model training. The second stage of hard sample selection further improved the performance to the SoTA level of $92.47\%$. This validates and confirms our proposal that information contained in hard samples is valuable for the model to learn a robust representation.

\vspace{-2mm}
\subsection{Visualization of Hard Anchor Hallucination}
\vspace{-1mm}

To demonstrate the effectiveness of our method, 
Fig.~\ref{fig:exp:tsne} presents the t-SNE visualization of our hallucination on the CIFAR-10 dataset with 40\% classification-based label noise in various training epochs. For simplicity, we limit the display to 25 hallucinated and 500 real samples for each class randomly sampled from $\mathcal{D}$. The colors of darker hues indicate the hallucinated anchors with pseudo-labels that match the corresponding lighter shades. Observe that the features of hallucinated anchors for each class align with the corresponding cluster of real features. This demonstrates that our hallucinated anchors can effectively mimic the desired hard samples with appropriate pseudo-labels, which can facilitate the subsequent hard sample selection for improved decision boundary training. We provide additional visualization on the hard anchors in the supplementary materials.

\section{Conclusions}
\vspace{-1mm}

In this paper, we present a novel framework to tackle the underestimation of hard samples in classic selection-based Noisy-Label Learning (NLL) methods. By leveraging easy samples to hallucinate the hard anchors, our approach captures crucial information from hard samples in the presence of instance-dependent noise. While we could not cover all the possible works, we compared with the most similar ones and demonstrated the effectiveness of our model on several benchmark datasets, achieving superior performance compared to state-of-the-art methods.
We believe that our work offers a fresh perspective on the significance of hard samples in training models under label noise, a factor frequently overlooked by conventional NLL methods. We show that leveraging the critical labeling information in clean hard samples can enhance the robustness of the decision boundary. Other domains may also benefit from our proposal, such as active learning, which also focuses on leveraging the information of the data effectively.

{\bf Limitations.}
Our framework identifies hard samples and corrects their labels through hard anchor hallucination, with the assumption that the selected easy feature set $\mathcal{S}_{e}$ (and hence the hallucinated anchor set $\mathcal{S}_{hal}$) span the class of interest.
As a result, the proposed hallucination process might not work well for highly imbalanced datasets.

{\bf Future work.} A thorough investigation and evaluation of the proposed framework on larger real-world datasets will preferably generate new insights to improve the current solution. We also plan to integrate the proposed framework into other domains beyond image classification to enhance the generalizability of our work.

\bibliographystyle{IEEEbib}
\bibliography{ICIP2024/ICIP2024_conference}

\end{document}